%% file: main.tex
\documentclass[10pt,twocolumn,letterpaper]{article}

\pdfoutput=1

\usepackage[pagenumbers]{cvpr} 

\input{preamble}

\definecolor{cvprblue}{rgb}{0.21,0.49,0.74}
\usepackage[pagebackref,breaklinks,colorlinks,citecolor=cvprblue]{hyperref}

\def\paperID{1581} 
\def\confName{CVPR}
\def\confYear{2024}

\title{Efficient Architecture Search via Bi-level Data Pruning}

\author{
Chongjun Tu\textsuperscript{\rm 1}, Peng Ye\textsuperscript{\rm 1}, Weihao Lin\textsuperscript{\rm 1}, Hancheng Ye\textsuperscript{\rm 1}, \\
Chong Yu\textsuperscript{\rm 1}, Tao Chen\textsuperscript{\rm 1}\thanks{Corresponding author}, Baopu Li,  Wanli Ouyang\textsuperscript{\rm 2}\\
\textsuperscript{\rm 1}Fudan University, \textsuperscript{\rm 2}Shanghai AI Laboratory\\
{\tt\small cjtu23@m.fudan.edu.cn}
}

\begin{document}
\maketitle

\input{sec/0_abstract}    
\input{sec/1_intro}

\input{sec/2_related}

\input{sec/3_methods}

\input{sec/4_experiments}

\input{sec/5_conclusion}
\input{sec/N_finalcopy}
{
    \small
    \bibliographystyle{ieeenat_fullname}
    \bibliography{main}
}


\end{document}

%% file: preamble.tex
\usepackage[dvipsnames]{xcolor}
\usepackage{graphics}
\usepackage{multicol}
\usepackage{multirow}
\usepackage{float}
\usepackage{tabularx}
\usepackage{tikz}
\usetikzlibrary{decorations}
\usepackage{booktabs}
\usepackage{diagbox}
\usepackage{algorithm}
\usepackage{algpseudocode}

\usepackage{colortbl,booktabs}

\definecolor{baselinecolor}{gray}{.9}

\DeclareMathOperator{\st}{s.t.}
\newcommand{\affiliations}1{

\pgfdeclaredecoration{varrow}{final}{
  \state{final}{
    \pgfpathmoveto{\pgfpointdecoratedpathlast}
    \pgfpathlineto{\pgfpoint{-2mm}{1mm}}
    \pgfpathlineto{\pgfpoint{-2mm}{-1mm}}
    \pgfpathclose
    \pgfusepath{fill}
  }
  \state{initial}[width=\pgfdecoratedinputsegmentremainingdistance]{
    \pgfpathmoveto{\pgfpointorigin}
    \pgfpathlineto{\pgfpoint{\pgfdecoratedinputsegmentremainingdistance}{0pt}}
    \pgfsetlinewidth{2mm}
    \pgfusepath{stroke}
  }
}

%% file: sec/0_abstract.tex
\begin{abstract}

Improving the efficiency of Neural Architecture Search (NAS) is a challenging but significant task that has received much attention.
Previous works mainly adopt the Differentiable Architecture Search (DARTS)
and improve its search strategies or modules to enhance search efficiency. 
Recently, some methods start considering data reduction for speedup, 
but they are not tightly coupled with the architecture search process, resulting in sub-optimal performance.
To this end, this work pioneers an exploration into the critical role of dataset characteristics for DARTS bi-level optimization, and then proposes a novel Bi-level Data Pruning (BDP) paradigm that targets the weights and architecture levels of DARTS to enhance efficiency from a data perspective. 
Specifically, we introduce a new progressive data pruning strategy that utilizes supernet prediction dynamics as the metric, to gradually prune unsuitable samples for DARTS during the search. An effective automatic class balance constraint is also integrated into BDP, to suppress potential class imbalances resulting from data-efficient algorithms. 
Comprehensive evaluations on the NAS-Bench-201 search space, DARTS search space, and MobileNet-like search space validate that BDP reduces search costs by over 50\% while achieving superior performance when applied to baseline DARTS. Besides, we demonstrate that BDP can harmoniously integrate with advanced DARTS variants, like PC-DARTS and $\beta$-DARTS, offering an approximately 2$\times$ speedup with minimal performance compromises.
\end{abstract}


%% file: sec/1_intro.tex
\section{Introduction}
\label{sec:intro}

\begin{figure}
    \centering
    \includegraphics[width=\linewidth]{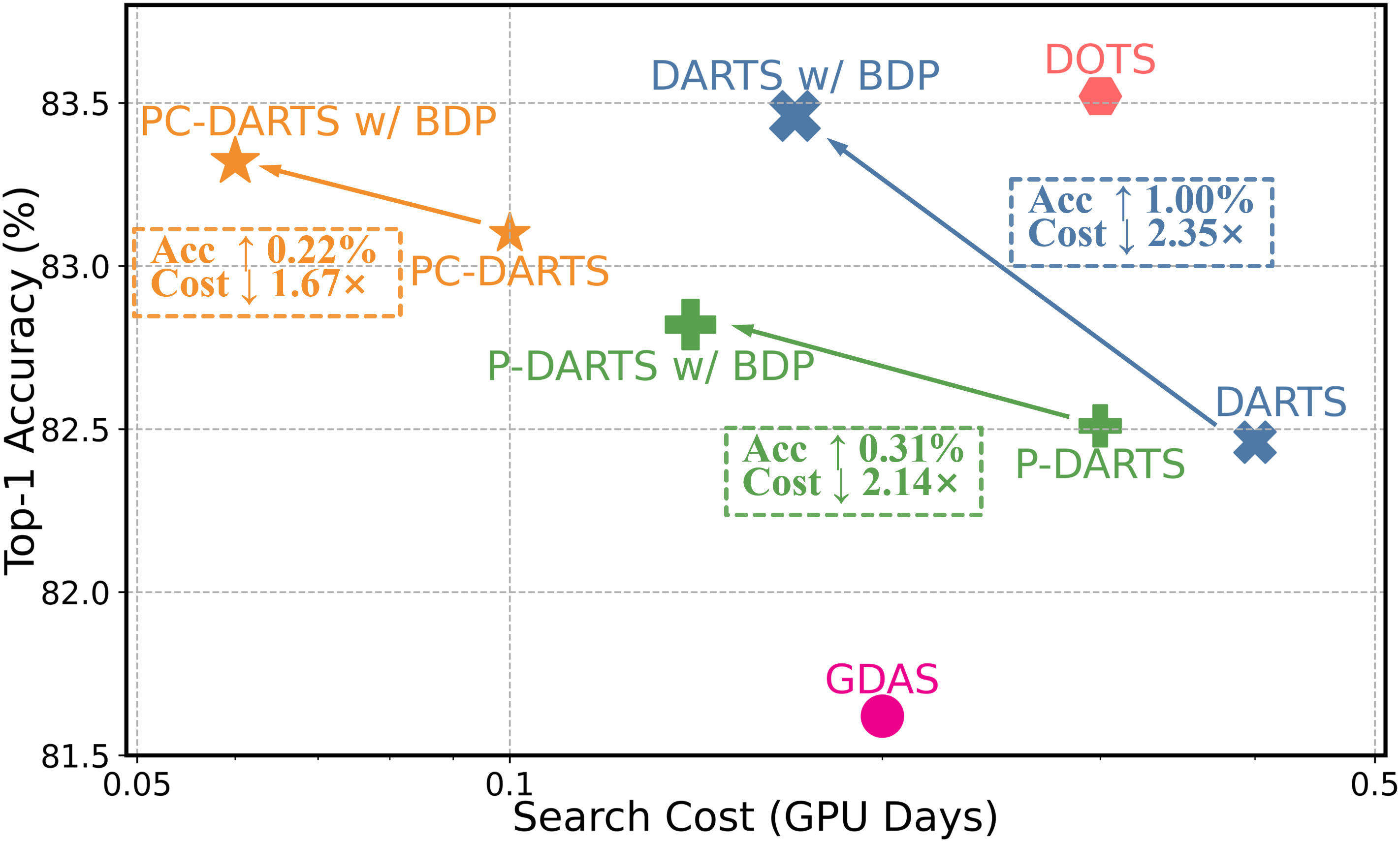}
    \vspace{-6mm}
    \caption{Comparison of search cost and top-1 accuracy for various DARTS algorithms on DARTS search space and CIFAR-100. Our proposed BDP provides evident accelerations for DARTS variants, achieving comparable or even higher performances.
    }
    \label{fig:enter-label}
    \vspace{-3mm}
\end{figure}

Neural architecture search (NAS) has shown superior performance 
across various tasks~\cite{chen2018searching,du2020spinenet,shen2022efficient,guo2022generalized,zoph2018learning,mei2023automatic,zhan2023a2s}. 
Among existing NAS methods, Differentiable Architecture Search (DARTS)~\cite{liu2018darts} and its variants~\cite{chu2020darts,chen2019progressive,xu2019pc,ye2022beta} stand out for their efficiency
in searching for an optimal architecture. 
For lower computational overhead, some works further put efforts on improving DARTS efficiency
by pruning candidates from the operation sets in the search space~\cite{chen2019progressive,ye2022efficient}, or developing new search strategies~\cite{cai2018proxylessnas,xu2019pc}. 
However, the search cost grows proportionally significant with larger scale of datasets, still posing a challenge for resource cost. This inspires researchers to consider boosting the efficiency of DARTS by strategically down-scaling the datasets.

Recent advances in data-efficient training, such as subset selection~\cite{killamsetty2021glister, zheng2022coverage, killamsetty2021grad}, have given rise to several works~\cite{na2021accelerating, shim2021core, prasad2022speeding} that aim to accelerate DARTS by reducing training data. They typically involve constructing a proxy or a subset of the dataset before the architecture search process.
However, these methods mostly decoupled the subset selection from the search process and overlooked architecture updates. They train the continuously varying architectures with invariant subsets~\cite{na2021accelerating,shim2021core}, which compromises the performance of the searched architectures.

\begin{table*}[htbp]
\vspace{-2mm}
\begin{multicols}{2}
\begin{table}[H]
\centering
\caption{Impact of different training and validation sets scale rates on DARTS performance, evaluated on NAS-Bench-201~\cite{dong2020bench}. The setting of baseline DARTS~\cite{liu2018darts} is marked in \colorbox{baselinecolor}{gray}. Here, T and V refer to the training set and the validation set, respectively.}
\label{tab: motivation1: split}
\vspace{-2mm}
\resizebox{0.9\linewidth}{!}{
\setlength{\tabcolsep}{1.5mm}
\begin{tabular}{ccccc}
\toprule
 & \multicolumn{2}{c}{CIFAR-10} & \multicolumn{2}{c}{CIFAR-100} \\ 
\cmidrule(lr){2-5} 
\raisebox{2pt}{\multirow{-2}{*}{T : V}} & valid  & test  & valid  & test  \\ 
\midrule
{1 : 9} & 39.77$\pm$0.00 & 54.30$\pm$0.00 & 15.03$\pm$0.00 & 15.61$\pm$0.00 \\
{3 : 7} & 39.77$\pm$0.00 & 54.30$\pm$0.00 & 15.03$\pm$0.00 & 15.61$\pm$0.00 \\
\rowcolor[gray]{0.9}{5 : 5} & 39.77$\pm$0.00 & 54.30$\pm$0.00 & 15.03$\pm$0.00 & 15.61$\pm$0.00 \\
{7 : 3} & 68.29$\pm$0.00 & 70.92$\pm$0.00 & 38.57$\pm$0.00 & 38.97$\pm$0.00 \\
{9 : 1} & 63.48$\pm$21.71 & 69.79$\pm$14.96 & 36.06$\pm$19.89 & 36.41$\pm$19.64 \\ 
\bottomrule
\end{tabular}
}
\end{table}

\begin{table}[H]
\centering
\caption{Effect of data distribution on DARTS performance for NAS-Bench-201~\cite{dong2020bench}. The top 50\% of samples with the lowest (denoted as \textbf{Low}) or highest (\textbf{High}) EL2N scores~\cite{paul2021deep} are discarded in the training (T) and validation (V) sets.}
\label{tab: motivation2: distribution}
\vspace{-1.8mm}
\resizebox{0.95\linewidth}{!}{
\setlength{\tabcolsep}{1.5mm}
\begin{tabular}{cccccc}
\toprule
\raisebox{-4pt}{\multirow{2}{*}{T}}  & \raisebox{-4pt}{\multirow{2}{*}{V}} &  \multicolumn{2}{c}{CIFAR-10} &
  \multicolumn{2}{c}{CIFAR-100} \\ 
\cmidrule(lr){3-6} 
 &   &   valid           & test            & valid           & test            \\
\midrule
\multicolumn{2}{c}{DARTS~\cite{liu2018darts}} & 39.77$\pm$0.00  & 54.30$\pm$0.00      & 15.03$\pm$0.00      & 15.61$\pm$0.00      \\ 
\cmidrule(lr){1-6} 
Low              & Low             & 39.77$\pm$0.00  & 54.30$\pm$0.00  & 15.03$\pm$0.00  & 15.61$\pm$0.00  \\
Low              & High            & 79.30$\pm$9.54  & 82.30$\pm$9.86  & 53.02$\pm$12.58 & 53.26$\pm$12.42 \\
High             & Low             & 51.58$\pm$20.46 & 63.06$\pm$15.17 & 25.86$\pm$18.75 & 26.38$\pm$18.66 \\
High             & High            & 56.37$\pm$28.75 & 67.21$\pm$22.36 & 33.08$\pm$31.26 & 33.60$\pm$31.17 \\ 
\bottomrule
\end{tabular}%
}
\end{table}
\end{multicols}
\vspace{-9mm}
\end{table*}

To tailor a data-efficient approach for DARTS, we first in-depth explore the impact of some dataset characteristics on the bi-level optimization (see ~\cref{sec:method: Preliminaries} for details of DARTS): 
\textit{\textbf{C1. Dataset scale.}} 
Recent findings suggest that the ideal dataset scale is correlated with the number of parameters to be optimized~\cite{sorscher2022beyond}. 
Therefore, the impact of training and validation set scales on DARTS is worth exploring.
\textit{\textbf{C2. Data distribution.}} 
Data distribution plays a significant role in training models. Many studies have explored this, categorizing samples as easy/hard~\cite{paul2021deep, sorscher2022beyond} or clear/ambiguous~\cite{swayamdipta2020dataset, he2023large} with different metrics.
These studies shed light on the relationship between the network and datasets and are suitable for our exploration purpose.

We conduct some exploratory experiments on NAS-Bench-201~\cite{dong2020bench} and present the results in ~\cref{tab: motivation1: split} and ~\cref{tab: motivation2: distribution}. 
For \textbf{\textit{C1.}}, we compare the results of different dataset splits with the typical equal split of DARTS in ~\cref{tab: motivation1: split} and find that uneven dataset splits can hopefully yield superior performances. 
For \textbf{\textit{C2.}}, we employ the widely applied EL2N metric~\cite{paul2021deep, attendu2023nlu, fayyaz2022bert, zayed2023deep} to change the data distributions of training and validation sets by pruning 50\% samples with highest/lowest EL2N scores. We can conclude from ~\cref{tab: motivation2: distribution} that different data distributions result in disparate performances. Interestingly, searching with pruned data can surpass the baseline by a large margin, emphasizing the significant impact of \textbf{\textit{C2.}} on DARTS.

Preliminary findings reveal the critical impact of dataset scales and distributions on the performance of DARTS bi-level optimization. 
However, current data-efficient DARTS methods~\cite{na2021accelerating,shim2021core,prasad2022speeding} strictly maintain identical scales and distributions for training and validation sets, which might not be ideal for DARTS optimization. 
To this end, we introduce Bi-level Data Pruning (BDP), a novel data-efficient paradigm for enhancing DARTS efficiency by progressively downscaling datasets. The \textit{bi-level} in BDP aligns with DARTS bi-level optimization, denoting the supernet weights level (training set) and architecture parameters level (validation set).
Specifically, BDP comprises a prediction-error-based metric termed Variance of Error (VoE), based on which we customize a data pruning strategy for DARTS.
This paradigm tackles the training dynamics of the supernet and adjusts the samples to be pruned in time. 
An automatic class balance constraint is also integrated into BDP to alleviate potential class imbalance issues studied in ~\cite{sorscher2022beyond} arising from data-efficient algorithms. 
We confirm the effectiveness of BDP on three commonly applied DARTS search spaces (NAS-Bench-201, DARTS, MobileNet-like) and provide the effectiveness analysis.


To summarize, our main contributions are as follows: 
\begin{enumerate}
    \item Through preliminary experiments on NAS-Bench-201 \cite{dong2020bench}, we reveal that the scales and data distributions of both training and validation sets significantly influence the bi-level optimization of DARTS, which has rarely been explored or discussed in previous studies.
    \item We propose Bi-level Data Pruning (BDP), a pioneering data-efficient paradigm deeply coupled with DARTS, to gradually downscale the dataset during the search process and boost DARTS efficiency and performance.
    Specifically, BDP involves a novel VoE metric, a progressive data pruning strategy, and an integrated automatic class balance constraint. 
    \item The proposed BDP is orthogonal to most methods of improving DARTS and can harmoniously integrate with different DARTS variants. 
    When applied to baseline DARTS~\cite{liu2018darts}, superior performance is obtained with over 50\% reduction of search cost. With advanced DARTS variants (\eg, PC-DARTS and $\beta$-DARTS), comparable performance can be achieved with about 2$\times$ speedup.
\end{enumerate}

%% file: sec/2_related.tex
\section{Related Work}
\label{sec:related}

\subsection{Efficient Architecture Search}
\textbf{Neural architecture search (NAS)} has achieved significant success, but early works~\cite{tan2019mnasnet,real2019regularized} suffered from unbearable overheads. 
Among various strategies to reduce costs~\cite{zhong2020blockqnn,real2019aging,cai2019once,zhang2020one}, one-shot architecture search has been favored due to the effective weight-sharing strategy.
Building upon this strategy, Differentiable Architecture Search (DARTS)~\cite{liu2018darts} introduced differentiable architecture parameters and a bi-level optimization technique to enhance the architecture search efficiency further. However, computational costs remain high. 
Efforts to accelerate DARTS focused on search strategy design~\cite{cai2018proxylessnas,xu2019pc} and search space shrinking~\cite{chen2019progressive,ye2022efficient}. 
Yet the search costs, scaling with dataset scales, suggest a promising and under-explored way to speed up DARTS from a data perspective.

\textbf{Data-efficient DARTS.} A few studies tried to improve DARTS efficiency using data-efficient algorithms. 
An early attempt~\cite{na2021accelerating} used an entropy-based method to select a proxy of the dataset before the search. 
Similarly, ~\cite{shim2021core} utilized the low-dimensional distribution of samples to derive a proxy. 
Concurrent work~\cite{yang2023revisiting} only retained a subset of the classes and randomly selected samples to form a highly condensed proxy.
However, the decoupling of dataset downscaling and architecture search exacerbates the performance of DARTS. 
Adaptive NAS~\cite{prasad2022speeding} updated the proxy dataset continuously based on GLISTER~\cite{killamsetty2021glister}. 
Nevertheless, it still suffered from compromised performances due to negligence of the distinctiveness of the training and validation sets in the bi-level optimization.
Besides, existing methods rarely consider class imbalance caused by dataset down-scaling~\cite{sorscher2022beyond}.
Compared to existing methods, our proposed paradigm deeply couples with the bi-level optimization of DARTS
and alleviates the potential class imbalance, which helps boost performances while providing evident acceleration. 


\subsection{Data-efficient Algorithms}
Data-efficient algorithms, such as subset selection and data pruning, are increasingly popular in deep learning for resource-constrained and high-overhead scenarios.
\textbf{Subset selection}
aims to identify a representative subset of the entire dataset. 
Different approaches have been developed, such as geometry-based~\cite{har2004coresets, mirzasoleiman2020coresets} and gradient-based~\cite{killamsetty2021grad}, to dissect the intrinsic characteristics of the dataset to select critical samples. 
\textbf{Data pruning}, on the other hand, focuses on removing redundant samples to improve training efficiency. It uses various metrics like EL2N~\cite{paul2021deep}, forgetting~\cite{toneva2018empirical}, memorization~\cite{feldman2020neural}, and dynamic uncertainty~\cite{he2023large} to score and rank the samples. Each metric then applies a corresponding data pruning strategy to determine and remove redundant samples. 
We discuss the challenges of applying data-efficient algorithms to DARTS in ~\cref{sec:method: Preliminaries}.

%% file: sec/3_methods.tex
\section{Method}
\label{method}

\subsection{Preliminaries}
\label{sec:method: Preliminaries}
\noindent{\textbf{Differentiable Architecture Search (DARTS)}} 
aims to find optimal cells to stack an entire architecture. These cells are represented as directed acyclic graphs (DAGs) with nodes as latent representations.
An edge between two nodes corresponds to a mixture of operations from the candidate set $\mathcal{O}$, where $\mathcal{O}_k$ denotes a single operation such as \textit{convolution}. 
DARTS introduces differentiable architecture parameters $\boldsymbol{\alpha}$ to weight all operations in $\mathcal{O}$ and constructs a hybrid operation $\bar{\mathcal{O}}$ on each edge:
\begin{equation}
\label{eq: mix ops}
\bar{\mathcal{O}}\left(x\right)=\sum_{k=1}^{\left|\mathcal{O}\right|}\beta_{k}\mathcal{O}_{k}\left(x\right),
\beta_{k}=\frac{\exp\left(\alpha_{k}\right)}{\sum_{i=1}^{|\mathcal{O}|}\exp\left(\alpha_{i}\right)},
\end{equation}
where $x$ is the input from a prior node, and $\boldsymbol{\beta}$ represents the \textit{softmax} normalized architecture parameters. 
DARTS then constructs a supernet that includes all architectures and solves the following bi-level optimization to search for the optimal architecture:
\begin{equation}
\label{eq: bi-level optimization}
\min_{\boldsymbol{\alpha}}\mathcal{L}_{\operatorname{val}}\left(\boldsymbol{w}^{*}\left(\boldsymbol{\alpha}\right),\boldsymbol{\alpha}\right),
\st \boldsymbol{w}^{*}\left(\boldsymbol{\alpha}\right)=\arg\min_{\boldsymbol{w}}\mathcal{L}_{\operatorname{train}}(\boldsymbol{w},\boldsymbol{\alpha}),
\end{equation}
where $\boldsymbol{w}$ is the weights w.r.t. the architecture derived from $\boldsymbol{\alpha}$. 
For ~\cref{eq: bi-level optimization}, DARTS equally divides the dataset into a training set for updating $\boldsymbol{w}$ and a validation set for $\boldsymbol{\alpha}$.
The final architecture is built by selecting operations with the highest $\boldsymbol{\alpha}$ on each edge at the end of the search process.

\begin{figure*}[htbp] 
    \centering 
    \includegraphics[width=0.8\linewidth]{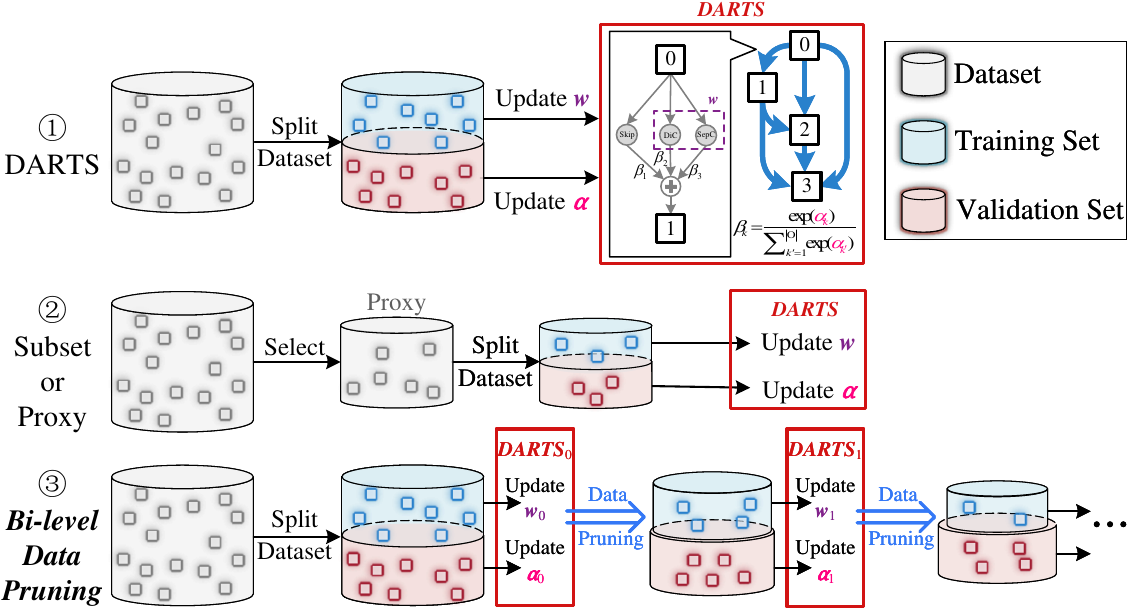} 
    \vspace{-2mm}
    \caption{
    Schematic comparisons between our proposed BDP paradigm and other data-efficient methods for DARTS: 
    \textbf{1)} Original DARTS divides the entire dataset equally into training and validation sets and performs bi-level optimization. 
    \textbf{2)} Existing methods use a specific strategy to select samples and construct a proxy of the entire dataset, which is then evenly split into training and validation sets for DARTS.
    \textbf{3)} Our BDP paradigm deeply couples with bi-level optimization and performs progressive data pruning separately on the training and validation sets. Thus, scales and data distributions of these two sets can be different during the architecture search process.
    }
    \label{fig:framework_compare}
    \vspace{-3mm}
\end{figure*}

\subsection{Bi-level Data Pruning (BDP) Paradigm}
\label{sec:method: BDP}
In this paper, we propose a bi-level data pruning (BDP) paradigm to enhance both the efficiency and performance of DARTS. 
BDP comprises a novel data pruning metric (VoE), a progressive data pruning strategy,
and an automatic class balance constraint. A comprehensive framework comparison between BDP and other methods is presented in ~\cref{fig:framework_compare}.

\subsubsection{Data-efficient Algorithms Review}
\label{sec: method limitations}
We briefly review data-efficient algorithms and discuss their challenges when applied to DARTS. 
Subset selection aims to approximate the entire dataset, which is not aligned with the best practices for DARTS, as evidenced by ~\cref{tab: motivation1: split} and ~\ref{tab: motivation2: distribution}.
Therefore, data pruning methods that can change the data distributions as necessary are more suitable for DARTS but face two significant limitations:
\textit{\textbf{L1. Additional overhead.}} Approaches like the forgetting~\cite{toneva2018empirical} and uncertainty~\cite{he2023large} scores require training statistics from full epochs, while EL2N~\cite{paul2021deep} needs multiple training runs, imposing an overhead that contradicts the efficiency goal;
\textit{\textbf{L2. One-shot data pruning.}} These methods prune substantial samples at once before searching, potentially discarding valuable samples for the search and failing to cope with architecture updates during the search process. 
We resolve these limitations by the design of the data pruning metric and strategy.

\subsubsection{Variance of Error (VoE) Metric}
To construct a data pruning metric suitable for DARTS, we take the EL2N metric~\cite{paul2021deep} as our foundation, chosen for its ability to evaluate samples without long training phases. 
The EL2N metric scores samples based on the model's prediction error.
Let $(x_i, y_i)$ denote a specific sample and its target. We compute the EL2N score for $(x_i, y_i)$ at the $t-$th training epoch with the network weights $w_t$.
The network's prediction probability for this sample is $p(w_t, x_i)$, normalized by \textit{softmax}. 
Then the EL2N score can be calculated:
\begin{equation}
\operatorname{EL2N}\left(x_{i}, y_{i},t\right)=\left\|p\left(\boldsymbol{w}_{t},x_{i}\right)-y_{i}\right\|_{2}, 
\end{equation}
where $ \left\| \cdot \right\|_{2}$ means the L2 norm. 
For the DARTS scenario, since the bi-level optimization updates $\boldsymbol{w}$ and $\boldsymbol{\alpha}$ with gradient descent (GD), the prediction probabilities of the supernet for each sample in the training and validation sets are effortlessly accessible. 
 
We then eliminate the additional overheads associated with sample evaluation (\textbf{\textit{L1}} in \cref{sec: method limitations}). ~\cite{paul2021deep} integrate the average of 5 runs to avoid the effect of weights initialization, thus incurring significant overhead. 
We take inspiration from the exponential moving average (EMA) technique and dynamic uncertainty~\cite{he2023large} to remove the dependence on initialization by taking statistics over time. Specifically, we propose a metric called VoE to score each sample by the variance of the supernet's prediction error. For brevity, we denote 
$e_t = \left\|p\left(\boldsymbol{w}_{t},x_{i}\right)-y_{i}\right\|_{2}$. Then we can calculate:
\begin{equation}
\operatorname{VoE}\left(x_i, y_i, t_0, \mathcal{T}\right) = \frac{1}{\mathcal{T}} \sum_{k=t_0}^{t_0+\mathcal{T}-1} \left(e_k - \bar{e}_k\right)^2,
\label{equ: VoE}
\end{equation}
where we refer to $\mathcal{T}$ as the data pruning interval as well as the temporal window for calculating VoE scores. 
With this metric, we can evaluate the architecture's prediction dynamics for each sample during the search with little overhead.

\subsubsection{Progressive Data Pruning Strategy}
\label{sec: method pruning strategy}
With the VoE metric, 
we customize a data pruning strategy to determine how to prune samples for DARTS.
Considering the continuous updates of the architecture parameters (\textbf{\textit{L2}} in \cref{sec: method limitations}), 
we propose to perform progressive data pruning, discarding small numbers of samples at intervals rather than all at once. 
This is similar to how ~\cite{chen2019progressive} and ~\cite{ye2022efficient} gradually narrow the candidate set of operations. 
Besides, as can be concluded from ~\cref{tab: motivation1: split} and ~\ref{tab: motivation2: distribution}, both the training and validation sets in the bi-level optimization impact the performance immensely and require separate consideration. Therefore, in BDP we prune samples from these two sets separately. 
Specifically, we prune samples with a $p_t\%$ ratio from the training set ($T$) and $p_v\%$ from the validation set ($V$) every $\mathcal{T}$ epochs. 

The remaining question is which samples to prune each time. We refer to this configuration as the \textit{data pruning criterion}, which is part of the data pruning strategy. To explore the best practice, we conduct an exploratory experiment on NAS-Bench-201~\cite{dong2020bench}, where five pruning ratio combinations and four criteria are considered. 
The results in~\cref{tab: criterion} demonstrate that the best criterion is to prune $p_t\%$ samples with the lowest VoE scores in the training set and $p_v\%$ samples with the highest VoE scores in the validation set.

\begin{table}[t]
\centering
\caption{
The effect of different data pruning criteria on the NAS-Bench-201 benchmark~\cite{dong2020bench}, where \textbf{Low} means pruning samples with lowest VoE scores from the dataset while \textbf{High} means the opposite.
Five pruning ratios $(p_t/p_v)\%=[(25/5), (20/10), (15/15), (10/20), (5/25)]$ are applied to each criterion and we record the average and the best results. }
\label{tab: criterion}
\vspace{-3mm}
\setlength{\tabcolsep}{1.8mm}
\resizebox{1.0\linewidth}{!}{
\begin{tabular}{cccccccc}
\toprule
\multicolumn{2}{c}{Criterion} & \multicolumn{2}{c}{CIFAR-10} & \multicolumn{2}{c}{CIFAR-100} & \multicolumn{2}{c}{Img.16-120}\\
\cmidrule(lr){1-2} \cmidrule(lr){3-8} 
{$T$}  & {$V$}    & {Avg.} & {Best} & {Avg.} & {Best} & {Avg.} & {Best} \\
\midrule
Low  & Low  & 54.30  & 90.25 & 25.36  & 64.34 & 16.32 & 16.32 \\
\textbf{Low}  & \textbf{High} & \textbf{81.41}  & \textbf{93.62} & \textbf{65.45}  & \textbf{72.99} & \textbf{32.19} & \textbf{44.57} \\
High & Low  & 57.62  & 70.92 & 46.25  & 63.52 & 16.74 & 18.41 \\
High & High & 66.77  & 90.36 & 15.61  & 15.61 & 22.77 & 38.59 \\
\bottomrule
\end{tabular}}
\vspace{-5mm}
\end{table}

\subsubsection{Automatic Class Balance Constraint} 
Findings of ~\cite{sorscher2022beyond} reveal that data-efficient algorithms may lead to or worsen class imbalance in the dataset, and severe data imbalance affects the performance. 
Our proposed BDP paradigm similarly faces this risk. 
Therefore, we propose a novel automatic class balance constraint and integrate it into BDP, which sets an upper limit for each class at each data pruning to avoid the class imbalance from becoming too severe. We follow~\cite{sorscher2022beyond} to define the class balance degree $b\in\left[0,1\right]$ of a dataset with $M$ classes as:
\begin{equation}
b = \frac{1}{P} \sum_{i=1}^{M-1} \sum_{j=i+1}^{M} \frac{\min (|class_i|, |class_j|)}{\max (|class_i|, |class_j|)}, \\
\end{equation}
where $P=\binom{M}{2}$ denotes the total number of class pairs.
$\left|class_i\right|$ is the number of samples in the $class_i$. From the definition, the higher $b$ is, the more balanced the classes are. 
We then construct a constraint for BDP: For $class_i$, the upper limit of the samples to be pruned is set to
\begin{equation}
\label{eq: constraint strength}
L_i=\frac{\left|class_i\right|}{N}, 
N=n\left(1-b^{2}\right)+1, 
\end{equation}
where $n$ denotes the total data pruning times. 
For a search process of $E$ epochs and data pruning interval $\mathcal{T}$, we can calculate that $n=E / \mathcal{T}$.
The constraint intensity $N \in \left[1,  n+1 \right)$ is a function of $b$, automatically restricting data pruning as the class balance degree of the current dataset declines. 
The design of $N$ involves the following considerations:
First, when $b=1$, the dataset is perfectly balanced: at this point $N=1$, data pruning can be performed according to $(p_t,p_v)$ freely; 
As data pruning proceeds and class imbalance emerges, $b$ gradually decreases: $N$ increases accordingly in time, automatically constraining the data pruning to alleviate the imbalance; 
Finally, if $b$ is close to 0, \ie, the data is entirely unbalanced: $N$ approaches $n$, resulting in a very conservative data pruning to avoid excessively few samples in some classes. 
We provide the comparison with other expressions of N in the \textbf{Supplementary Material}.

\begin{algorithm}
\caption{BDP Paradigm for DARTS}
\label{alg: algorithm of BDP}
\begin{algorithmic}[1]
\Require Training set $T$ and Validation set $V$. Data pruning interval $\mathcal{T}$. Data pruning ratios $(p_t, p_v)$ for $(T, V)$.

\Ensure The optimal searched architecture.

\Statex \hspace{-10pt}$\blacklozenge$ \textbf{Initializations}
\Statex Construct a supernet and initialize the architecture parameters $\alpha$ and supernet weights $w$.

\Statex \hspace{-10pt}$\blacklozenge$ \textbf{Architecture search process}
\For {$epoch$ in epochs}
    \For {$batch_i$ in $T$ and $batch_j$ in $V$}
        \State Update $\alpha$ with $batch_j$;
        \State Update $w$ with $batch_i$;
        \State Record the prediction error $e_t$ of each sample;
    \EndFor
    \If{$epoch$ \% $\mathcal{T}$ == 0 }
        \State Compute the VoE score of remaining samples;
        \State Set constraints for each class in $T$ and $V$;
        \State Prune $p_t\%$ samples with the lowest scores of $T$
        \Statex \hspace{30pt}and $p_v\%$ samples with the highest scores of $V$;
    \EndIf
\EndFor

\State Derive the searched architecture based on $\alpha$.
\end{algorithmic}
\end{algorithm}

%% file: sec/4_experiments.tex
\section{Experiments}
\label{experiments}

In this section, we present the results of applying the BDP paradigm to various DARTS variants (DARTS, P-DARTS~\cite{chen2019progressive}, PC-DARTS\cite{xu2019pc}, and $\beta$-DARTS\cite{ye2022beta}) on different datasets (CIFAR-10, CIFAR-100, ImageNet) and different search spaces (NAS-Bench-201, DARTS, MobileNet-like). The comprehensive results demonstrate the efficiency and effectiveness of BDP. 
The search cost is reported on a single NVIDIA Tesla V100 unless otherwise stated. 
We further demonstrate the effect of class balance constraints in the ablation study. 
Moreover, we also conduct experiments to verify why and how BDP works, bringing insights into the importance of data for DARTS. The overall algorithm of BDP is shown in ~\cref{alg: algorithm of BDP}.

\subsection{Experiments on NAS-Bench-201 Search Space}
\textbf{Settings: }NAS-Bench-201 is a popular benchmark that enables quick performance evaluation of NAS algorithms. It comprises a fixed search space containing 15,625 possible cell structures, with each consisting of 4 nodes and 5 operations. Meanwhile, the benchmark provides the performance of each architecture (called genotype) on three datasets (CIFAR-10, CIFAR-100, and ImageNet16-120). We apply BDP to the baseline DARTS~\cite{liu2018darts} and advanced $\beta$-DARTS~\cite{ye2022beta} and conduct various experiments on NAS-Bench-201. Each experiment is repeated 3 times with different random seeds. The performance of searched architectures are queried by their genotypes. The settings of the search process are consistent with the baseline DARTS~\cite{liu2018darts}. 

\begin{table*}[htbp] 
\centering
\caption{Comparisons with state-of-the-art (SOTA) methods and existing data-efficient DARTS methods on NAS-Bench-201 benchmark~\cite{dong2020bench} are shown in the upper block and the bottom block respectively. 
Note that $^\dagger$ and $^\ddagger$ indicate that the search is conducted on the CIFAR-10 and CIFAR-100 datasets, respectively. Our reported results are averaged on 4 independent runs of experiments.}
\vspace{-1mm}
\label{Main: NAS-Bench-201}
\resizebox{1.0\linewidth}{!}{
\setlength{\tabcolsep}{2.4mm}
\begin{tabular}{ccccccccc}
\toprule
\multirow{2}{*}{Methods} &
  \multirow{2}{*}{\begin{tabular}[c]{@{}c@{}}Cost\\ (hours)\end{tabular}} &
  \multirow{2}{*}{\begin{tabular}[c]{@{}c@{}}Data\\ (\%)\end{tabular}} &
  \multicolumn{2}{c}{CIFAR-10} &
  \multicolumn{2}{c}{CIFAR-100} &
  \multicolumn{2}{c}{ImageNet16-120} \\ 
  \cmidrule(lr){4-9}
    &   &   & valid    & test    & valid    & test    & valid    & test    \\ 
    \midrule
DARTS(1st)~\cite{liu2018darts}    & 3.2   &  100  & 39.77$\pm$0.00 & 54.30$\pm$0.00 & 15.03$\pm$0.00  & 15.61$\pm$0.00  & 16.43$\pm$0.00 & 16.32$\pm$0.00 \\
DARTS(2nd)~\cite{liu2018darts}    & 10.2    &  100  & 39.77$\pm$0.00 & 54.30$\pm$0.00 & 15.03$\pm$0.00  & 15.61$\pm$0.00  & 16.43$\pm$0.00 & 16.32$\pm$0.00 \\
GDAS~\cite{dong2019searching}    & 8.7    &  100  & 89.89$\pm$0.08 & 93.61$\pm$0.09 & 71.34$\pm$0.04  & 70.70$\pm$0.30  & 41.59$\pm$1.33 & 41.71$\pm$0.98 \\
SNAS~\cite{xie2018snas}    & -    &  100   & 90.10$\pm$1.04 & 92.77$\pm$0.83 & 69.69$\pm$2.39  & 69.34$\pm$1.98  & 42.84$\pm$1.79 & 43.16$\pm$2.64 \\
DSNAS~\cite{hu2020dsnas}    & -    &   100  & 89.66$\pm$0.29 & 93.08$\pm$0.13 & 30.87$\pm$16.40 & 31.01$\pm$16.38 & 40.61$\pm$0.09 & 41.07$\pm$0.09 \\
PC-DARTS~\cite{xu2019pc}    & -    &   100  & 89.96$\pm$0.15 & 93.41$\pm$0.30 & 67.12$\pm$0.39  & 67.48$\pm$0.89  & 40.83$\pm$0.08 & 41.31$\pm$0.22 \\
iDARTS~\cite{zhang2021idarts}    & -    &    100   & 89.86$\pm$0.60 & 93.58$\pm$0.32 & 70.57$\pm$0.24  & 70.83$\pm$0.48  & 40.38$\pm$0.59 & 40.89$\pm$0.68 \\
DARTS-~\cite{chu2020darts}    & 3.2    &  100   & 91.03$\pm$0.44 & 93.80$\pm$0.40 & 71.36$\pm$1.51  & 71.53$\pm$1.51  & 44.87$\pm$1.46 & 45.12$\pm$0.82 \\
CDARTS~\cite{yu2022cyclic}    & -    &    100  & 91.12$\pm$0.44 & 94.02$\pm$0.31 & 72.12$\pm$1.23  & 71.92$\pm$1.30  & 45.09$\pm$0.61 & 45.51$\pm$0.72 \\
$\beta$-DARTS$^\dagger$~\cite{ye2022beta}    & 3.2    &   100 & 91.55$\pm$0.00 & 94.36$\pm$0.00 & 73.49$\pm$0.00  & 73.51$\pm$0.00  & 46.37$\pm$0.00 & 46.34$\pm$0.00 \\ 
DARTS w/ BDP$^\dagger$    & 1.6    & 100$\Rightarrow$4   & 90.98$\pm$0.24 & 93.79$\pm$0.39	& 71.33$\pm$1.46 & 71.74$\pm$2.17 & 44.36$\pm$1.09 & 44.67$\pm$1.33   \\
DARTS w/ BDP$^\ddagger$    & 1.9    & 100$\Rightarrow$14   & 90.97$\pm$0.13 & 93.76$\pm$0.28	& 71.61$\pm$0.74 & 71.64$\pm$1.30 & 44.84$\pm$0.56 & 45.12$\pm$0.39   \\
$\beta$-DARTS w/ BDP$^\dagger$  & 1.6    & 100$\Rightarrow$4   & 91.57$\pm$0.03 & 94.36$\pm$0.01	& 73.24$\pm$0.43 & 73.41$\pm$0.17 & 46.10$\pm$0.47 & 46.46$\pm$0.21   \\
$\beta$-DARTS w/ BDP$^\ddagger$ & 1.9    & 100$\Rightarrow$14   & 91.55$\pm$0.00 & 94.36$\pm$0.00 & 73.49$\pm$0.00 & 73.51$\pm$0.00 & 46.37$\pm$0.00  & 46.34$\pm$0.00   \\ 
\cmidrule(lr){1-9}
Adaptive-NAS~\cite{prasad2022speeding}    & 0.8 & 10   & - & 92.22$\pm$1.76   & -  & 64.27$\pm$3.37& - & 36.10$\pm$6.96 \\
DARTS w/ BDP          & 0.7  & 100$\Rightarrow$1  & 89.64$\pm$2.35  & 92.58$\pm$2.04  & 69.81$\pm$2.95  & 70.05$\pm$3.13  & 43.44$\pm$2.25  & 43.41$\pm$2.89    \\
$\beta$-DARTS w/ BDP  & 0.7  & 100$\Rightarrow$1  & 91.59$\pm$0.03  & 94.37$\pm$0.01  & 73.00$\pm$0.43  & 73.32$\pm$0.17  & 45.83$\pm$0.47  & 46.59$\pm$0.21    \\
\cmidrule(lr){1-9}
optimal    &    &    & 91.61    & 94.37    & 73.49    & 73.51    & 46.77    & 47.31    \\ 
\bottomrule
\end{tabular}}
\vspace{-4mm}
\end{table*}

\textbf{Results: }
In~\cref{Main: NAS-Bench-201}, we compare the results obtained using different methods. It can be observed from the upper block that applying BDP to the baseline DARTS can not only reduce search costs but also boost performance. DARTS w/ BDP progressively reduces the data to only 4$\%$/14$\%$ of the entire dataset and completes the search on CIFAR-10 and CIFAR-100 in only 1.6h and 1.9h, respectively. It also produces architectures that are comparable to advanced methods that aim to improve the performance or robustness of DARTS.
In addition, $\beta$-DARTS w/ BDP achieves a 2$\times$ speedup while still being able to search for nearly optimal architectures on the NAS-Bench-201 space. 
We note that the overall performance of $\beta$-DARTS w/ BDP$^\dagger$ on CIFAR-100 has slightly declined due to an experiment that finds the optimal architecture on CIFAR-10 (94.37), whose performance on CIFAR-100 is 73.22, which is lower than the optimal accuracy of CIFAR-100 (73.51). This resulted in a lower average performance on CIFAR-100. 
Nevertheless, the proposed BDP is still effective for $\beta$-DARTS$^\dagger$ as it is conducted on CIFAR-10.

\begin{figure*}[htbp] 
	\centering 
	\includegraphics[width=0.9\linewidth]{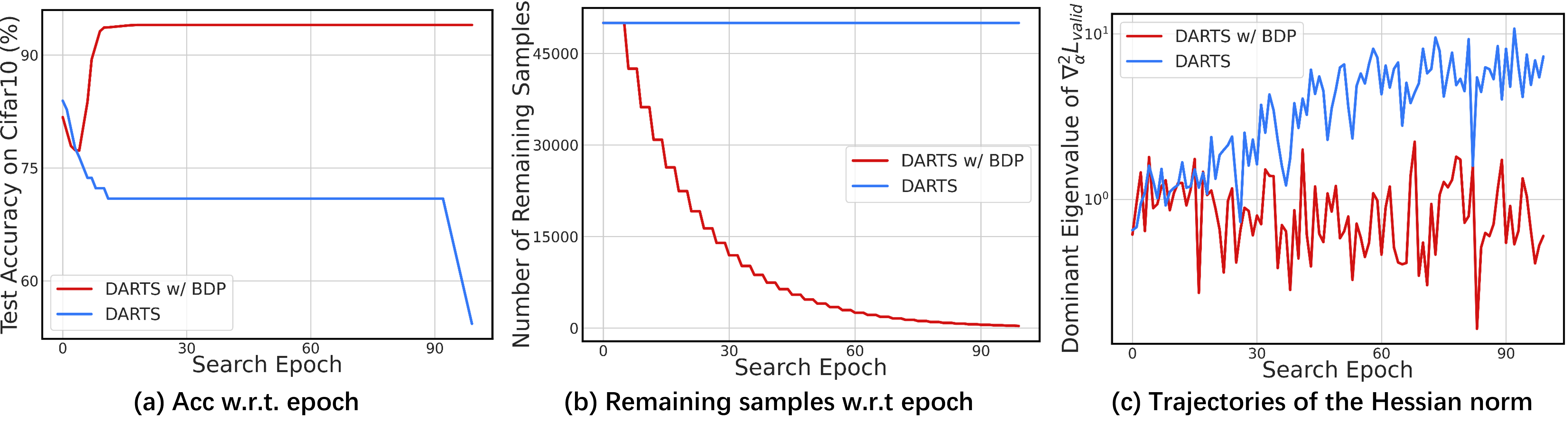} 
        \vspace{-4mm}
	\caption{Comprehensive comparison of the search process of baseline DARTS and DARTS w/ BDP on NAS-Bench-201 benchmark~\cite{dong2020bench}: (a) Test accuracy on CIFAR-10; (b) Number of remaining samples; (c) The trajectories of the dominant eigenvalues of $\nabla ^{2} _{\alpha} L_{valid}$. For baseline DARTS, the performance collapses, the number of samples remains unchanged, and the dominant eigenvalue persistently increases. For DARTS w/ BDP, the performance gradually increases, the number of samples decreases, and the dominant eigenvalue oscillates down.}
	\label{fig: bench_change}
 \vspace{-1mm}
\end{figure*}

As shown in the bottom block of ~\cref{Main: NAS-Bench-201}, DARTS w/ BDP can obtain better results than Adaptive-NAS~\cite{prasad2022speeding} when the search time is aligned, which is based on more advanced DARTS-PT~\cite{wang2021rethinking}. Meanwhile, $\beta$-DARTS w/ BDP can maintain almost optimal performances.
We also show in (a) and (b) of ~\cref{fig: bench_change} the change of accuracy and dataset scales when performing DARTS w/ BDP on the NAS-Bench-201 search space. It can be observed that DARTS w/ BDP achieves relatively good performance early in the training process and does not experience severe performance degradation, which further illustrates the effectiveness of our proposed BDP.

\subsection{Experiments on DARTS Search Space}
\textbf{Settings: }The DARTS search space~\cite{liu2018darts} can provide high-performance architectures on multiple datasets and is also a widely used search space for evaluating NAS algorithms. This space contains two types of cells: normal cells and reduction cells. Each cell has 4 intermediate nodes and 14 edges, and each edge has 8 candidate operations. 
Similar to NAS-Bench-201, we apply BDP to the baseline DARTS~\cite{liu2018darts} and advanced variants including P-DARTS~\cite{chen2019progressive}, PC-DARTS~\cite{xu2019pc}, and $\beta$-DARTS~\cite{ye2022beta}, following settings in~\cite{liu2018darts}. 
We conduct these experiments with 3 different random seeds on the CIFAR-10 and CIFAR-100 datasets and report the best results.
We also follow ~\cite{xu2019pc} in transferring the architectures searched on CIFAR to the ImageNet dataset and searching directly on the ImageNet.

\textbf{Results: }
On the left side of ~\cref{tab:Main: DARTS search space}, we show the results on the CIFAR datasets.
For baseline DARTS, our proposed BDP reduces the search time to less than half of the baseline and surprisingly improves its performance from 97.00\% and 82.46\% to 97.46\% and 83.46\% on CIFAR-10 and CIFAR-100, comparable with the SOTA methods. 
For other variants, BDP can also provide an approximate 2 times speedup with minimal performance compromise, even improving performance on the CIFAR-100 dataset for P-DARTS and PC-DARTS. Compared with the proxy-based DARTS~\cite{na2021accelerating}, applying our proposed BDP takes a longer time to search, but it dramatically mitigates performance decline and finds architectures that exceed or are close to the original performance, which cannot be achieved by using a proxy. We have also aligned with ~\cite{na2021accelerating} on DARTS and find that BDP can find architectures of better performance. 

We compare the results on the ImageNet dataset on the right side of ~\cref{tab:Main: DARTS search space}. 
The architectures searched by DARTS variants on the CIFAR dataset can perform well when transferred to ImageNet, demonstrating excellent generalization when applying BDP. 
In addition, we also directly search on ImageNet with the setting of PC-DARTS w/ BDP. We reduced the search cost from 3.8 to 1.2 days with a slight performance improvement.

\begin{table*}[htbp] 
\begin{center}
\caption{Comprehensive comparison of different methods on DARTS search space on CIFAR (left) and ImageNet (right) datasets. For CIFAR, results in the top and mid blocks are reported by~\cite{ye2022beta}. Results in the bottom block show the results of applying BDP or proxy~\cite{na2021accelerating} to DARTS variants. 
For ImageNet, in the blocks from top to bottom, we consider the comparison of other search spaces (\textit{italicized}), cross-domains (denoted as CD.), transferring, and applying our proposed BDP paradigm, respectively. 
$^\dagger$, $^\ddagger$ and $^\ast$ represent that the search is conducted on the CIFAR-10, CIFAR-100, and ImageNet datasets, respectively.
}
\vspace{-2mm}
\label{tab:Main: DARTS search space}
\resizebox{\linewidth}{!}{
\begin{tabular}{lcccccllccccc}
\cmidrule[0.08em]{1-6} \cmidrule[0.08em]{8-12}
\multirow{2}{*}{Method} & \multirow{2}{*}{\begin{tabular}[c]{@{}c@{}}GPU\\ (Days)\end{tabular}} & \multicolumn{2}{c}{CIFAR-10} & \multicolumn{2}{c}{CIFAR-100} &  & \multirow{2}{*}{Method}    & \multirow{2}{*}{\begin{tabular}[c]{@{}c@{}}GPU\\ (Days)\end{tabular}} & \multirow{2}{*}{\begin{tabular}[c]{@{}c@{}}Params\\ (M)\end{tabular}}  & \multirow{2}{*}{\begin{tabular}[c]{@{}c@{}}Top1\\ (\%)\end{tabular}} & \multirow{2}{*}{\begin{tabular}[c]{@{}c@{}}Top5\\ (\%)\end{tabular}} \\ \cline{3-6}
    &    & Params(M)    & Acc(\%)    & Params(M)    & Acc(\%)    &  &    &    &    &    &    &    \\ 
    \cline{1-6} \cline{8-12} 
NASNet-A~\cite{zoph2018learning}    & 2000    & 3.3    & 97.35    & 3.3    & 83.18    &  &  \textit{MnasNet-92}$^\ast$~\cite{tan2019mnasnet}    & 1667    & 4.4       & 74.8    & 92.0    \\
DARTS(1st)~\cite{liu2018darts}    & 0.4    & 3.4    & 97.00$\pm$0.14    & 3.4    & 82.46    &  & \textit{FairDARTS}$^\ast$~\cite{chu2020fair}    & 3    & 4.3        & 75.6    & 92.6    \\
DARTS(2nd)~\cite{liu2018darts}    & 1    & 3.3    & 97.24$\pm$0.09    & -    & -    &  & PC-DARTS$^\ast$~\cite{xu2019pc}    & 3.8    & 5.3        & 75.8    & 92.7  \\
SNAS~\cite{xie2018snas}    & 1.5    & 2.8    & 97.15$\pm$0.02    & 2.8    & 82.45    &  & DOTS$^\ast$~\cite{gu2021dots}    & 1.3    & 5.3        & 76.0    & 92.8    \\
GDAS~\cite{dong2019searching}    & 0.2    & 3.4    & 97.07    & 3.4    & 81.62    &  & \textit{DARTS}-$^\ast$~\cite{chu2020darts}    & 4.5    & 4.9        & 76.2    & 93.0    \\ 
\cline{8-12} 
P-DARTS~\cite{chen2019progressive}    & 0.3    & 3.4    & 97.50    & 3.6    & 82.51    &  & \multicolumn{1}{c}{AdaptNAS-S(CD.)~\cite{li2020adapting}} & 1.8 & 5.0  & 74.7  & 92.2 \\
PC-DARTS~\cite{xu2019pc}    & 0.1    & 3.6    & 97.43$\pm$0.07    & 3.6    & 83.10    &  &  \multicolumn{1}{c}{AdaptNAS-C(CD.)~\cite{li2020adapting}} & 2.0    & 5.3        & 75.8    & 92.6    \\ 
\cline{1-6} \cline{8-12} 
P-DARTS~\cite{chen2019progressive}    & 0.3    & 3.3$\pm$0.21    & 97.19$\pm$0.14    & -    & -    & & AmoebaNet-C$^\dagger$~\cite{real2019regularized}    & 3150    & 6.4        & 75.7    & 92.4    \\
R-DARTS(L2)~\cite{zela2019understanding}    & 1.6    & -    & 97.05$\pm$0.21    & -    & 81.99$\pm$0.26    &  & SNAS$^\dagger$~\cite{xie2018snas}    & 1.5    & 4.3       & 72.7    & 90.8    \\
SDARTS-ADV~\cite{chen2020stabilizing}    & 1.3    & 3.3    & 97.39$\pm$0.02    & -    & -    &  & P-DARTS$^\ddagger$~\cite{chen2019progressive}    & 0.3    & 5.1        & 75.3    & 92.5    \\
DOTS~\cite{gu2021dots}    & 0.3    & 3.5    & 97.51$\pm$0.06    & 4.1    & 83.52$\pm$0.13    &  & SDARTS-ADV$^\dagger$~\cite{chen2020stabilizing}    & 1.3    & 5.4        & 74.8    & 92.2    \\
DARTS+PT~\cite{wang2021rethinking}    & 0.8    & 3.0    & 97.39$\pm$0.08    & -    & -    &  & DOTS$^\dagger$~\cite{gu2021dots}    & 0.3    & 5.2        & 75.7    & 92.6    \\
DARTS-~\cite{chu2020darts}    & 0.4    & 3.5$\pm$0.13    & 97.41$\pm$0.08    & 3.4    & 82.49$\pm$0.25    &  &  DARTS+PT$^\dagger$~\cite{wang2021rethinking}    & 0.8    & 4.6        & 74.5    & 92.0    \\
$\beta$-DARTS$^\dagger$~\cite{ye2022beta}    & 0.4    & 3.78$\pm$0.08    & 97.49$\pm$0.07    & 3.83$\pm$0.08    & 83.48$\pm$0.03    &  &  $\beta$-DARTS$^\ddagger$~\cite{ye2022beta}    & 0.4    & 5.4    & 75.8  & 92.9    \\
$\beta$-DARTS$^\ddagger$~\cite{ye2022beta}    & 0.4    & 3.75$\pm$0.15    & 97.47$\pm$0.08    & 3.80$\pm$0.15    & 83.76$\pm$0.22    &  &  $\beta$-DARTS$^\dagger$~\cite{ye2022beta}    & 0.4   & 5.5     & 76.1 & 93.0    \\
\cmidrule(lr){1-6} \cmidrule(lr){8-12}

DARTS w/ BDP    & 0.17  & 2.94 & 97.46$^\dagger$ & 3.60  & 83.46$^\ddagger$   & & DARTS w/ BDP$^\dagger$  & 0.2 & 5.5 & 75.7 & 92.8  \\
$\beta$-DARTS w/ BDP  & 0.22  &  3.69	& 97.47$^\dagger$ 	&  3.74	& 83.33$^\ddagger$   &  &  $\beta$-DARTS w/ BDP$^\ddagger$ & 0.2 & 5.2 & 75.4 & 92.4  \\
P-DARTS w/ BDP  & 0.14  &  3.64	& 97.21$^\dagger$ &  4.01	& 82.82$^\ddagger$ &   &$\beta$-DARTS w/ BDP$^\dagger$ & 0.2 & 5.3 & 75.7 & 92.6  \\
PC-DARTS w/ BDP & 0.06 &  3.65  & 97.30$^\dagger$ &  3.70 & 83.32$^\ddagger$ &  & P-DARTS w/ BDP$^\ddagger$ & 0.2 & 6.1 & 76.0 & 92.7  \\
\cline{1-6}

DARTS w/ proxy~\cite{na2021accelerating}    & 0.03  &    -    & 97.06    &    -    &    -    &  &  PC-DARTS w/ BDP$^\ddagger$ & 0.2 & 5.5 & 75.7 & 92.5  \\
DARTS w/ BDP      & 0.04  &    4.35 & 97.18    &    -    &    -    &  &  PC-DARTS w/ BDP$^\ast$ & 1.2 & 5.5  & 75.8 & 92.8 \\

\cmidrule[0.08em]{1-6} \cmidrule[0.08em]{8-12}
\end{tabular}}
\vspace{-6mm}
\end{center}
\end{table*}

\subsection{Experiments on MobileNet-like Search Space}
\textbf{Settings: }
The MobileNet search space is structured differently from cell-based search spaces like NAS-Bench-201 and DARTS search space. It is a chain-structured space often used to evaluate various NAS methodologies. This space comprises 21 layers, each with an inverted bottleneck MBConv with expansion rate options of 3 and 6. It also has kernel dimension candidates of 3, 5, 7, and a skip connection that can be used to explore models of different depths. 
During the search phase, we follow ~\cite{ye2023beta} to randomly sample 12.5\% of the ImageNet dataset and divide them into a 4:1 ratio for the initial training and validation sets for the DARTS bi-level optimization. The search process is carried out for 30 epochs with a batch size of 640. 
To ensure fair comparisons, our settings for retraining and evaluation are consistent with those in ~\cite{li2020block} and ~\cite{ye2023beta}. The search cost is reported on 8 NVIDIA Tesla V100s.

\textbf{Results: }
The efficiency and effectiveness of our proposed BDP on MobileNet-like Search Space is shown in ~\cref{tab: mobilenet}. 
This table does not include the performance of the baseline DARTS because the searched architecture is dominated by skip connections, thus performing poorly. 
However, DARTS w/ BDP can search for architectures with a Top-1 accuracy of 76.7\% with only the search cost of 0.3 days. 
In parallel, we find that the BDP paradigm reduces the search cost of $\beta-$DARTS by more than two times, \ie, from 0.8days to 0.3days, at the cost of only a 0.4\% reduction in Top-1 accuracy and a 0.1\% reduction in Top-5 accuracy.

\begin{table}[t]
\centering
\caption{Results on the MobileNet search space. $^\ast$ in the upper block denotes using the DARTS search space, and the methods in the bottom block are directly searched on ImageNet. 
}
\label{tab: mobilenet}
\vspace{-2mm}
\resizebox{1.0\linewidth}{!}{
\setlength{\tabcolsep}{2mm}
\begin{tabular}{lcccccc}
\toprule
\multirow{2}{*}{Method}  & \multirow{2}{*}{\begin{tabular}[c]{@{}c@{}}GPU\\ (Days)\end{tabular}} & \multirow{2}{*}{\begin{tabular}[c]{@{}c@{}}FLOPs\\ (M)\end{tabular}}   & \multirow{2}{*}{\begin{tabular}[c]{@{}c@{}}Params\\ (M)\end{tabular}}  & \multirow{2}{*}{\begin{tabular}[c]{@{}c@{}}Top1\\ (\%)\end{tabular}} & \multirow{2}{*}{\begin{tabular}[c]{@{}c@{}}Top5\\ (\%)\end{tabular}} & \multirow{2}{*}{\begin{tabular}[c]{@{}c@{}}Proxy\\ data\end{tabular}} \\
\\
\midrule
PC-DARTS$^\ast$~\cite{xu2019pc}              & 3.8   & 597   & 5.3 & 75.8 & 92.7 & 12.5\% \\
ER-PC-DARTS$^\ast$~\cite{xu2021partially}             & 2.8   & 578   & 5.1 & 75.9 & 92.8 & 12.5\% \\
\midrule
MobileNetV3~\cite{koonce2021mobilenetv3}             & 3000  & 219   & 5.4 & 75.2 & 92.2 & 100\% \\
MixNet-M~\cite{tan2019mixconv}                & 3000  & 360   & 5.0 & 77.0 & 93.3 & 100\% \\
EfficientNet B0~\cite{tan2019efficientnet}         & 3000  & 390   & 5.3 & 76.3 & 93.2 & 100\% \\
MoGA-A~\cite{chu2020moga}                  & 12    & 304   & 5.1 & 75.9 & 92.8 & 100\% \\
OFA~\cite{cai2019once}                     & 55    & 230   & {-} & 76.9 & {-}  & 100\% \\
DNA-b~\cite{li2020block}                   & 24.6  & 406   & 4.9 & 77.5 & 93.3 & 100\% \\
NoisyDARTS-A~\cite{chu2020noisy}            & 12    & 449   & 5.5 & 77.9 & 94.0 & 100\% \\
DARTS-~\cite{chu2020darts}                  & 4.5   & 470   & 5.5 & 77.8 & 93.9 & 100\% \\
$\beta$-DARTS~\cite{ye2023beta}           & 0.8   & 476   & 4.8 & 78.1 & 93.8 & 12.5\% \\

DARTS w/ BDP            & 0.3   & 356   & 4.2 & 76.7 & 92.9 & 12.5\% \\
$\beta$-DARTS w/BDP     & 0.3   & 486   & 4.8 & 77.7 & 93.7 & 12.5\% \\
\bottomrule
\end{tabular}}
\vspace{-2mm}
\end{table}

\subsection{Ablation Study}
\label{sec: ablation}
To investigate the influence of class balance constraints on the proposed BDP paradigm, we conduct comparative experiments by applying BDP with and without class balance constraints to DARTS and $\beta$-DARTS on the NAS-Bench-201 benchmark. The experimental results are shown in ~\cref{tab:ablation: balance}. 
The findings indicate that the performance of DARTS w/ BDP drops significantly without the class balance constraint. In addition, even though $\beta$-DARTS is a robust algorithm, its performance decreases without the class balance constraint. Therefore, it is clear that the proposed class balance constraint is necessary and effective for applying BDP to DARTS.

\begin{table}[t]
\centering
\caption{Ablation study of the class balance constraints. We apply BDP with (without) class balance constraints to DARTS and $\beta$-DARTS on the NAS-Bench-201 search space~\cite{dong2020bench}. For each setting, we conducted 3 independent experiments on CIFAR-10 and CIFAR-100 with different random seeds.}
\vspace{-2mm}
\label{tab:ablation: balance}
\resizebox{1.0\linewidth}{!}{
\setlength{\tabcolsep}{1.5mm}
\begin{tabular}{cccccc}
\toprule
\multirow{2}{*}{Methods} 
& \multirow{2}{*}{balance} & \multicolumn{2}{c}{CIFAR-10} & \multicolumn{2}{c}{CIFAR-100} \\ 
\cmidrule(lr){3-6} 
&       & valid & test & valid & test \\ 
\midrule
DARTS w/ BDP      & \textbf{$\surd$} & 90.98$\pm$0.24 & 93.79$\pm$0.39 & 71.61$\pm$0.74 & 71.64$\pm$1.30    \\
DARTS w/ BDP      &        & 90.68$\pm$0.37 & 93.66$\pm$0.52 & 67.27$\pm$3.03 & 66.43$\pm$3.44    \\
$\beta$-DARTS w/ BDP & \textbf{$\surd$} & 91.57$\pm$0.03 & 94.36$\pm$0.01 & 73.49$\pm$0.00 & 73.51$\pm$0.00    \\
$\beta$-DARTS w/ BDP &       & 91.55$\pm$0.00 & 94.36$\pm$0.00 & 71.14$\pm$4.08 & 71.46$\pm$3.54    \\
\bottomrule
\end{tabular}}
\vspace{-2mm}
\end{table}

\subsection{Effectiveness Analysis}
\label{sec: effectiveness analysis}
\begin{figure}[t]
\centering 
\includegraphics[width=1.0\linewidth]{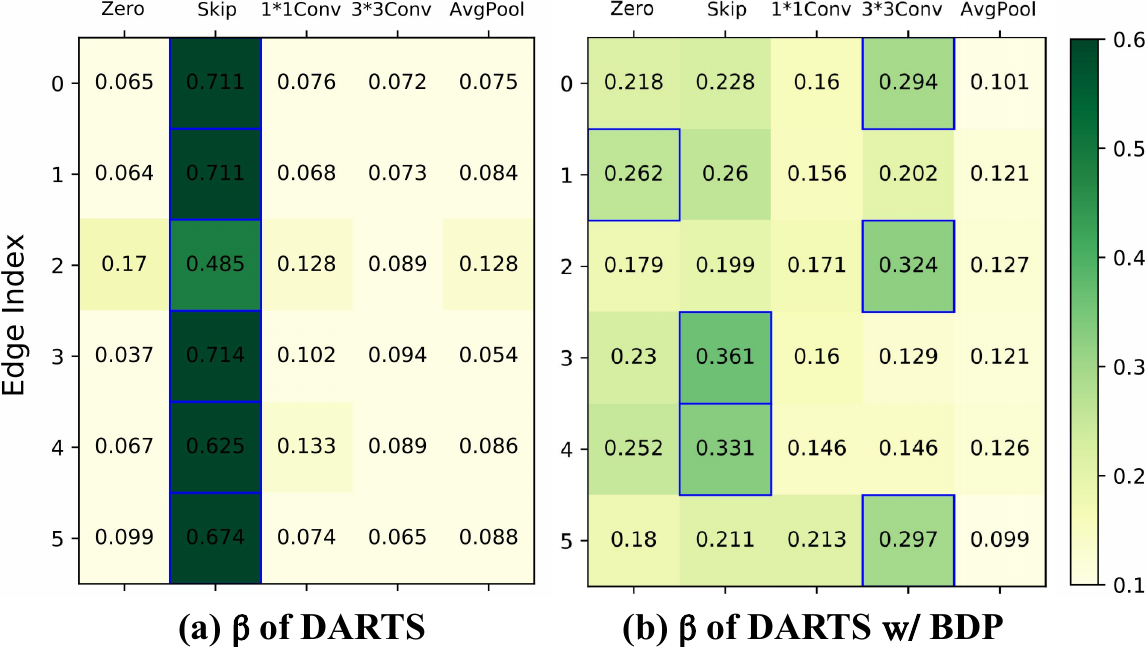} 
\vspace{-6mm}
\caption{Architecture parameters comparison of DARTS and DARTS w/ BDP after the search. The experiments are conducted on NAS-Bench-201~\cite{dong2020bench}. 
The vertical axis indicates the edge index, the horizontal axis indicates the five candidate operations and the number in each square shows the operation weight. 
The selected operation for each edge in the final architecture is marked with a rectangle. In the DARTS results, the weight of the skip connection dominates, while DARTS w/ BDP alleviates this problem.}
\label{fig: heatmap}
\vspace{-2mm}
\end{figure}

The proposed BDP paradigm can achieve comparable or higher performance despite using reduced training samples for DARTS. This seems counter-intuitive, as the sample reduction typically leads to a performance decline. 
We try to analyze the effectiveness of BDP from two aspects. 
Firstly, a typical issue in DARTS is the dominance of parameter-free operations,
significantly affecting the performance~\cite{zela2019understanding,ye2022beta}. 
BDP suppresses this issue as shown by the architecture parameters visualization at the end of baseline DARTS and DARTS w/ BDP in ~\cref{fig: heatmap}. 

Besides, we analyze the effectiveness of BDP from the perspective of the Hessian matrix $\nabla^2_\alpha L_{val}\left(w, \alpha\right)$ in the search process, which is closely related to the quality of DARTS solutions~\cite{zela2019understanding}. 
We provide a brief theoretical proof building upon ~\cite{chen2020stabilizing}. 
Let us define the pair $(w^\dagger, \alpha^\dagger)$ as the optimal solution of ~\cref{eq: bi-level optimization} and assume $\nabla_\alpha L_{val}(w^\dagger, \alpha^\dagger)=0$, which satisfies the optimality conditions. 
After the search, we obtain a discrete architecture $\hat{\alpha}$ by mapping $\alpha^\dagger$ to a discrete domain. The validation loss of $\hat{\alpha}$ can be calculated through Taylor expansion as:
\begin{equation}
L_{val}(w^\dagger, \hat{\alpha}) = L_{val}(w^\dagger, \alpha^\dagger) + \frac{1}{2}(\hat{\alpha} - \alpha^\dagger)^\top \tilde{H} (\hat{\alpha} - \alpha^\dagger),
\end{equation}
where $\tilde{H}$ denotes the cumulative Hessian matrix from $\alpha^\dagger$ to $\hat{\alpha}$. 
Assuming the Hessian matrix to remain consistent locally, the performance decline of deriving $\hat{\alpha}$ from $\alpha^\dagger$ with $w^\dagger$ can be approximated with the bound $B = \left\|\nabla_{\alpha}^2 L_{val}(w^\dagger, \alpha^\dagger)\right\|\left\|\hat{\alpha} - \alpha^\dagger\right\|^2$. 
In the retraining stage, where the weights are optimized, the validation loss tends to be lower than $L_{val}(w^\dagger, \hat{\alpha})$. Consequently, the performance of the retrained $\hat{\alpha}$ is also bounded by $B$. 
This confirms the Hessian norm's close relation to the quality and stability of DARTS solutions.
A large dominant eigenvalue of the Hessian matrix indicates potential performance drops during the architecture derivation.
Accordingly, we compare the dominant eigenvalue trajectories of $\nabla^2_\alpha L_{valid}$ during the search process between baseline DARTS and DARTS w/ BDP, as illustrated in (c) of ~\cref{fig: bench_change}. Interestingly, BDP can avoid the continuous rise in the eigenvalue of DARTS, which helps explain its effectiveness.

%% file: sec/5_conclusion.tex
\section{Conclusion}
In this paper, we explore the impact of training and validation sets on the bi-level optimization of DARTS, which is rarely discussed in previous works. 
Based on these observations, we introduce a novel data pruning paradigm called bi-level data pruning (BDP) that improves the efficiency and performance of DARTS. Comprehensive experiments on various search spaces and datasets indicate the effectiveness of BDP, which provides new insights into the relationship between DARTS and the datasets. 



%% file: sec/N_finalcopy.tex
%% file: main.bbl
\begin{thebibliography}{58}
\providecommand{\natexlab}[1]{#1}
\providecommand{\url}[1]{\texttt{#1}}
\expandafter\ifx\csname urlstyle\endcsname\relax
  \providecommand{\doi}[1]{doi: #1}\else
  \providecommand{\doi}{doi: \begingroup \urlstyle{rm}\Url}\fi

\bibitem[Attendu and Corbeil(2023)]{attendu2023nlu}
Jean-Michel Attendu and Jean-Philippe Corbeil.
\newblock Nlu on data diets: Dynamic data subset selection for nlp classification tasks.
\newblock \emph{arXiv preprint arXiv:2306.03208}, 2023.

\bibitem[Cai et~al.(2018)Cai, Zhu, and Han]{cai2018proxylessnas}
Han Cai, Ligeng Zhu, and Song Han.
\newblock Proxylessnas: Direct neural architecture search on target task and hardware.
\newblock \emph{arXiv preprint arXiv:1812.00332}, 2018.

\bibitem[Cai et~al.(2019)Cai, Gan, Wang, Zhang, and Han]{cai2019once}
Han Cai, Chuang Gan, Tianzhe Wang, Zhekai Zhang, and Song Han.
\newblock Once-for-all: Train one network and specialize it for efficient deployment.
\newblock \emph{arXiv preprint arXiv:1908.09791}, 2019.

\bibitem[Chen et~al.(2018)Chen, Collins, Zhu, Papandreou, Zoph, Schroff, Adam, and Shlens]{chen2018searching}
Liang-Chieh Chen, Maxwell Collins, Yukun Zhu, George Papandreou, Barret Zoph, Florian Schroff, Hartwig Adam, and Jon Shlens.
\newblock Searching for efficient multi-scale architectures for dense image prediction.
\newblock \emph{Advances in neural information processing systems}, 31, 2018.

\bibitem[Chen and Hsieh(2020)]{chen2020stabilizing}
Xiangning Chen and Cho-Jui Hsieh.
\newblock Stabilizing differentiable architecture search via perturbation-based regularization.
\newblock In \emph{International conference on machine learning}, pages 1554--1565. PMLR, 2020.

\bibitem[Chen et~al.(2019)Chen, Xie, Wu, and Tian]{chen2019progressive}
Xin Chen, Lingxi Xie, Jun Wu, and Qi Tian.
\newblock Progressive differentiable architecture search: Bridging the depth gap between search and evaluation.
\newblock In \emph{Proceedings of the IEEE/CVF international conference on computer vision}, pages 1294--1303, 2019.

\bibitem[Chu and Zhang(2020)]{chu2020noisy}
Xiangxiang Chu and Bo Zhang.
\newblock Noisy differentiable architecture search.
\newblock \emph{arXiv preprint arXiv:2005.03566}, 2020.

\bibitem[Chu et~al.(2020{\natexlab{a}})Chu, Wang, Zhang, Lu, Wei, and Yan]{chu2020darts}
Xiangxiang Chu, Xiaoxing Wang, Bo Zhang, Shun Lu, Xiaolin Wei, and Junchi Yan.
\newblock Darts-: robustly stepping out of performance collapse without indicators.
\newblock \emph{arXiv preprint arXiv:2009.01027}, 2020{\natexlab{a}}.

\bibitem[Chu et~al.(2020{\natexlab{b}})Chu, Zhang, and Xu]{chu2020moga}
Xiangxiang Chu, Bo Zhang, and Ruijun Xu.
\newblock Moga: Searching beyond mobilenetv3.
\newblock In \emph{ICASSP 2020-2020 IEEE International Conference on Acoustics, Speech and Signal Processing (ICASSP)}, pages 4042--4046. IEEE, 2020{\natexlab{b}}.

\bibitem[Chu et~al.(2020{\natexlab{c}})Chu, Zhou, Zhang, and Li]{chu2020fair}
Xiangxiang Chu, Tianbao Zhou, Bo Zhang, and Jixiang Li.
\newblock Fair darts: Eliminating unfair advantages in differentiable architecture search.
\newblock In \emph{European conference on computer vision}, pages 465--480. Springer, 2020{\natexlab{c}}.

\bibitem[Dong and Yang(2019)]{dong2019searching}
Xuanyi Dong and Yi Yang.
\newblock Searching for a robust neural architecture in four gpu hours.
\newblock In \emph{Proceedings of the IEEE/CVF Conference on Computer Vision and Pattern Recognition}, pages 1761--1770, 2019.

\bibitem[Dong and Yang(2020)]{dong2020bench}
Xuanyi Dong and Yi Yang.
\newblock Nas-bench-201: Extending the scope of reproducible neural architecture search.
\newblock \emph{arXiv preprint arXiv:2001.00326}, 2020.

\bibitem[Du et~al.(2020)Du, Lin, Jin, Ghiasi, Tan, Cui, Le, and Song]{du2020spinenet}
Xianzhi Du, Tsung-Yi Lin, Pengchong Jin, Golnaz Ghiasi, Mingxing Tan, Yin Cui, Quoc~V Le, and Xiaodan Song.
\newblock Spinenet: Learning scale-permuted backbone for recognition and localization.
\newblock In \emph{Proceedings of the IEEE/CVF conference on computer vision and pattern recognition}, pages 11592--11601, 2020.

\bibitem[Fayyaz et~al.(2022)Fayyaz, Aghazadeh, Modarressi, Pilehvar, Yaghoobzadeh, and Kahou]{fayyaz2022bert}
Mohsen Fayyaz, Ehsan Aghazadeh, Ali Modarressi, Mohammad~Taher Pilehvar, Yadollah Yaghoobzadeh, and Samira~Ebrahimi Kahou.
\newblock Bert on a data diet: Finding important examples by gradient-based pruning.
\newblock \emph{arXiv preprint arXiv:2211.05610}, 2022.

\bibitem[Feldman and Zhang(2020)]{feldman2020neural}
Vitaly Feldman and Chiyuan Zhang.
\newblock What neural networks memorize and why: Discovering the long tail via influence estimation.
\newblock \emph{Advances in Neural Information Processing Systems}, 33:\penalty0 2881--2891, 2020.

\bibitem[Gu et~al.(2021)Gu, Wang, Liu, Yang, Wu, Lu, and Cheng]{gu2021dots}
Yu-Chao Gu, Li-Juan Wang, Yun Liu, Yi Yang, Yu-Huan Wu, Shao-Ping Lu, and Ming-Ming Cheng.
\newblock Dots: Decoupling operation and topology in differentiable architecture search.
\newblock In \emph{Proceedings of the IEEE/CVF Conference on Computer Vision and Pattern Recognition}, pages 12311--12320, 2021.

\bibitem[Guo et~al.(2022)Guo, Chen, He, Liu, Xu, Ye, and Chen]{guo2022generalized}
Bicheng Guo, Tao Chen, Shibo He, Haoyu Liu, Lilin Xu, Peng Ye, and Jiming Chen.
\newblock Generalized global ranking-aware neural architecture ranker for efficient image classifier search.
\newblock In \emph{Proceedings of the 30th ACM International Conference on Multimedia}, pages 3730--3741, 2022.

\bibitem[Har-Peled and Mazumdar(2004)]{har2004coresets}
Sariel Har-Peled and Soham Mazumdar.
\newblock On coresets for k-means and k-median clustering.
\newblock In \emph{Proceedings of the thirty-sixth annual ACM symposium on Theory of computing}, pages 291--300, 2004.

\bibitem[He et~al.(2023)He, Yang, Huang, and Zhao]{he2023large}
Muyang He, Shuo Yang, Tiejun Huang, and Bo Zhao.
\newblock Large-scale dataset pruning with dynamic uncertainty.
\newblock \emph{arXiv preprint arXiv:2306.05175}, 2023.

\bibitem[Hu et~al.(2020)Hu, Xie, Zheng, Liu, Shi, Liu, and Lin]{hu2020dsnas}
Shoukang Hu, Sirui Xie, Hehui Zheng, Chunxiao Liu, Jianping Shi, Xunying Liu, and Dahua Lin.
\newblock Dsnas: Direct neural architecture search without parameter retraining.
\newblock In \emph{Proceedings of the IEEE/CVF Conference on Computer Vision and Pattern Recognition}, pages 12084--12092, 2020.

\bibitem[Killamsetty et~al.(2021{\natexlab{a}})Killamsetty, Durga, Ramakrishnan, De, and Iyer]{killamsetty2021grad}
Krishnateja Killamsetty, Sivasubramanian Durga, Ganesh Ramakrishnan, Abir De, and Rishabh Iyer.
\newblock Grad-match: Gradient matching based data subset selection for efficient deep model training.
\newblock In \emph{International Conference on Machine Learning}, pages 5464--5474. PMLR, 2021{\natexlab{a}}.

\bibitem[Killamsetty et~al.(2021{\natexlab{b}})Killamsetty, Sivasubramanian, Ramakrishnan, and Iyer]{killamsetty2021glister}
Krishnateja Killamsetty, Durga Sivasubramanian, Ganesh Ramakrishnan, and Rishabh Iyer.
\newblock Glister: Generalization based data subset selection for efficient and robust learning.
\newblock In \emph{Proceedings of the AAAI Conference on Artificial Intelligence}, pages 8110--8118, 2021{\natexlab{b}}.

\bibitem[Koonce and Koonce(2021)]{koonce2021mobilenetv3}
Brett Koonce and Brett Koonce.
\newblock Mobilenetv3.
\newblock \emph{Convolutional Neural Networks with Swift for Tensorflow: Image Recognition and Dataset Categorization}, pages 125--144, 2021.

\bibitem[Li et~al.(2020{\natexlab{a}})Li, Peng, Yuan, Wang, Liang, Lin, and Chang]{li2020block}
Changlin Li, Jiefeng Peng, Liuchun Yuan, Guangrun Wang, Xiaodan Liang, Liang Lin, and Xiaojun Chang.
\newblock Block-wisely supervised neural architecture search with knowledge distillation.
\newblock In \emph{Proceedings of the IEEE/CVF Conference on Computer Vision and Pattern Recognition}, pages 1989--1998, 2020{\natexlab{a}}.

\bibitem[Li et~al.(2020{\natexlab{b}})Li, Yang, Wang, and Xu]{li2020adapting}
Yanxi Li, Zhaohui Yang, Yunhe Wang, and Chang Xu.
\newblock Adapting neural architectures between domains.
\newblock \emph{Advances in neural information processing systems}, 33:\penalty0 789--798, 2020{\natexlab{b}}.

\bibitem[Liu et~al.(2018)Liu, Simonyan, and Yang]{liu2018darts}
Hanxiao Liu, Karen Simonyan, and Yiming Yang.
\newblock Darts: Differentiable architecture search.
\newblock \emph{arXiv preprint arXiv:1806.09055}, 2018.

\bibitem[Mei et~al.(2023)Mei, Ye, Ye, Li, Guo, Chen, and Ouyang]{mei2023automatic}
Zhen Mei, Peng Ye, Hancheng Ye, Baopu Li, Jinyang Guo, Tao Chen, and Wanli Ouyang.
\newblock Automatic loss function search for adversarial unsupervised domain adaptation.
\newblock \emph{IEEE Transactions on Circuits and Systems for Video Technology}, 2023.

\bibitem[Mirzasoleiman et~al.(2020)Mirzasoleiman, Bilmes, and Leskovec]{mirzasoleiman2020coresets}
Baharan Mirzasoleiman, Jeff Bilmes, and Jure Leskovec.
\newblock Coresets for data-efficient training of machine learning models.
\newblock In \emph{International Conference on Machine Learning}, pages 6950--6960. PMLR, 2020.

\bibitem[Na et~al.(2021)Na, Mok, Choe, and Yoon]{na2021accelerating}
Byunggook Na, Jisoo Mok, Hyeokjun Choe, and Sungroh Yoon.
\newblock Accelerating neural architecture search via proxy data.
\newblock \emph{arXiv preprint arXiv:2106.04784}, 2021.

\bibitem[Paul et~al.(2021)Paul, Ganguli, and Dziugaite]{paul2021deep}
Mansheej Paul, Surya Ganguli, and Gintare~Karolina Dziugaite.
\newblock Deep learning on a data diet: Finding important examples early in training.
\newblock \emph{Advances in Neural Information Processing Systems}, 34:\penalty0 20596--20607, 2021.

\bibitem[Prasad et~al.(2022)Prasad, White, Jain, Nayak, Iyer, and Ramakrishnan]{prasad2022speeding}
Vishak Prasad, Colin White, Paarth Jain, Sibasis Nayak, Rishabh~K Iyer, and Ganesh Ramakrishnan.
\newblock Speeding up nas with adaptive subset selection.
\newblock In \emph{First Conference on Automated Machine Learning (Late-Breaking Workshop)}, 2022.

\bibitem[Real et~al.(2019{\natexlab{a}})Real, Aggarwal, Huang, and Le]{real2019aging}
Esteban Real, Alok Aggarwal, Yanping Huang, and Quoc~V Le.
\newblock Aging evolution for image classifier architecture search.
\newblock In \emph{AAAI conference on artificial intelligence}, page~2, 2019{\natexlab{a}}.

\bibitem[Real et~al.(2019{\natexlab{b}})Real, Aggarwal, Huang, and Le]{real2019regularized}
Esteban Real, Alok Aggarwal, Yanping Huang, and Quoc~V Le.
\newblock Regularized evolution for image classifier architecture search.
\newblock In \emph{Proceedings of the aaai conference on artificial intelligence}, pages 4780--4789, 2019{\natexlab{b}}.

\bibitem[Shen et~al.(2022)Shen, Khodak, and Talwalkar]{shen2022efficient}
Junhong Shen, Misha Khodak, and Ameet Talwalkar.
\newblock Efficient architecture search for diverse tasks.
\newblock \emph{Advances in Neural Information Processing Systems}, 35:\penalty0 16151--16164, 2022.

\bibitem[Shim et~al.(2021)Shim, Kong, and Kang]{shim2021core}
Jae-hun Shim, Kyeongbo Kong, and Suk-Ju Kang.
\newblock Core-set sampling for efficient neural architecture search.
\newblock \emph{arXiv preprint arXiv:2107.06869}, 2021.

\bibitem[Sorscher et~al.(2022)Sorscher, Geirhos, Shekhar, Ganguli, and Morcos]{sorscher2022beyond}
Ben Sorscher, Robert Geirhos, Shashank Shekhar, Surya Ganguli, and Ari Morcos.
\newblock Beyond neural scaling laws: beating power law scaling via data pruning.
\newblock \emph{Advances in Neural Information Processing Systems}, 35:\penalty0 19523--19536, 2022.

\bibitem[Swayamdipta et~al.(2020)Swayamdipta, Schwartz, Lourie, Wang, Hajishirzi, Smith, and Choi]{swayamdipta2020dataset}
Swabha Swayamdipta, Roy Schwartz, Nicholas Lourie, Yizhong Wang, Hannaneh Hajishirzi, Noah~A Smith, and Yejin Choi.
\newblock Dataset cartography: Mapping and diagnosing datasets with training dynamics.
\newblock \emph{arXiv preprint arXiv:2009.10795}, 2020.

\bibitem[Tan and Le(2019{\natexlab{a}})]{tan2019efficientnet}
Mingxing Tan and Quoc Le.
\newblock Efficientnet: Rethinking model scaling for convolutional neural networks.
\newblock In \emph{International conference on machine learning}, pages 6105--6114. PMLR, 2019{\natexlab{a}}.

\bibitem[Tan and Le(2019{\natexlab{b}})]{tan2019mixconv}
Mingxing Tan and Quoc~V Le.
\newblock Mixconv: Mixed depthwise convolutional kernels.
\newblock \emph{arXiv preprint arXiv:1907.09595}, 2019{\natexlab{b}}.

\bibitem[Tan et~al.(2019)Tan, Chen, Pang, Vasudevan, Sandler, Howard, and Le]{tan2019mnasnet}
Mingxing Tan, Bo Chen, Ruoming Pang, Vijay Vasudevan, Mark Sandler, Andrew Howard, and Quoc~V Le.
\newblock Mnasnet: Platform-aware neural architecture search for mobile.
\newblock In \emph{Proceedings of the IEEE/CVF conference on computer vision and pattern recognition}, pages 2820--2828, 2019.

\bibitem[Toneva et~al.(2018)Toneva, Sordoni, Combes, Trischler, Bengio, and Gordon]{toneva2018empirical}
Mariya Toneva, Alessandro Sordoni, Remi Tachet~des Combes, Adam Trischler, Yoshua Bengio, and Geoffrey~J Gordon.
\newblock An empirical study of example forgetting during deep neural network learning.
\newblock \emph{arXiv preprint arXiv:1812.05159}, 2018.

\bibitem[Wang et~al.(2021)Wang, Cheng, Chen, Tang, and Hsieh]{wang2021rethinking}
Ruochen Wang, Minhao Cheng, Xiangning Chen, Xiaocheng Tang, and Cho-Jui Hsieh.
\newblock Rethinking architecture selection in differentiable nas.
\newblock \emph{arXiv preprint arXiv:2108.04392}, 2021.

\bibitem[Xie et~al.(2018)Xie, Zheng, Liu, and Lin]{xie2018snas}
Sirui Xie, Hehui Zheng, Chunxiao Liu, and Liang Lin.
\newblock Snas: stochastic neural architecture search.
\newblock \emph{arXiv preprint arXiv:1812.09926}, 2018.

\bibitem[Xu et~al.(2019)Xu, Xie, Zhang, Chen, Qi, Tian, and Xiong]{xu2019pc}
Yuhui Xu, Lingxi Xie, Xiaopeng Zhang, Xin Chen, Guo-Jun Qi, Qi Tian, and Hongkai Xiong.
\newblock Pc-darts: Partial channel connections for memory-efficient architecture search.
\newblock \emph{arXiv preprint arXiv:1907.05737}, 2019.

\bibitem[Xu et~al.(2021)Xu, Xie, Dai, Zhang, Chen, Qi, Xiong, and Tian]{xu2021partially}
Yuhui Xu, Lingxi Xie, Wenrui Dai, Xiaopeng Zhang, Xin Chen, Guo-Jun Qi, Hongkai Xiong, and Qi Tian.
\newblock Partially-connected neural architecture search for reduced computational redundancy.
\newblock \emph{IEEE Transactions on Pattern Analysis and Machine Intelligence}, 43\penalty0 (9):\penalty0 2953--2970, 2021.

\bibitem[Yang et~al.(2023)Yang, Yang, Jin, and Chen]{yang2023revisiting}
Taojiannan Yang, Linjie Yang, Xiaojie Jin, and Chen Chen.
\newblock Revisiting training-free nas metrics: An efficient training-based method.
\newblock In \emph{Proceedings of the IEEE/CVF Winter Conference on Applications of Computer Vision}, pages 4751--4760, 2023.

\bibitem[Ye et~al.(2022{\natexlab{a}})Ye, Li, Chen, Fan, Mei, Lin, Zuo, Chi, and Ouyang]{ye2022efficient}
Peng Ye, Baopu Li, Tao Chen, Jiayuan Fan, Zhen Mei, Chen Lin, Chongyan Zuo, Qinghua Chi, and Wanli Ouyang.
\newblock Efficient joint-dimensional search with solution space regularization for real-time semantic segmentation.
\newblock \emph{International Journal of Computer Vision}, 130\penalty0 (11):\penalty0 2674--2694, 2022{\natexlab{a}}.

\bibitem[Ye et~al.(2022{\natexlab{b}})Ye, Li, Li, Chen, Fan, and Ouyang]{ye2022beta}
Peng Ye, Baopu Li, Yikang Li, Tao Chen, Jiayuan Fan, and Wanli Ouyang.
\newblock $\beta$-darts: Beta-decay regularization for differentiable architecture search.
\newblock In \emph{2022 IEEE/CVF Conference on Computer Vision and Pattern Recognition (CVPR)}, pages 10864--10873. IEEE, 2022{\natexlab{b}}.

\bibitem[Ye et~al.(2023)Ye, He, Li, Chen, Bai, and Ouyang]{ye2023beta}
Peng Ye, Tong He, Baopu Li, Tao Chen, Lei Bai, and Wanli Ouyang.
\newblock $\beta $-darts++: Bi-level regularization for proxy-robust differentiable architecture search.
\newblock \emph{arXiv preprint arXiv:2301.06393}, 2023.

\bibitem[Yu et~al.(2022)Yu, Peng, Huang, Fu, Du, Wang, and Ling]{yu2022cyclic}
Hongyuan Yu, Houwen Peng, Yan Huang, Jianlong Fu, Hao Du, Liang Wang, and Haibin Ling.
\newblock Cyclic differentiable architecture search.
\newblock \emph{IEEE Transactions on Pattern Analysis and Machine Intelligence}, 45\penalty0 (1):\penalty0 211--228, 2022.

\bibitem[Zayed et~al.(2023)Zayed, Parthasarathi, Mordido, Palangi, Shabanian, and Chandar]{zayed2023deep}
Abdelrahman Zayed, Prasanna Parthasarathi, Gon{\c{c}}alo Mordido, Hamid Palangi, Samira Shabanian, and Sarath Chandar.
\newblock Deep learning on a healthy data diet: Finding important examples for fairness.
\newblock In \emph{Proceedings of the AAAI Conference on Artificial Intelligence}, pages 14593--14601, 2023.

\bibitem[Zela et~al.(2019)Zela, Elsken, Saikia, Marrakchi, Brox, and Hutter]{zela2019understanding}
Arber Zela, Thomas Elsken, Tonmoy Saikia, Yassine Marrakchi, Thomas Brox, and Frank Hutter.
\newblock Understanding and robustifying differentiable architecture search.
\newblock \emph{arXiv preprint arXiv:1909.09656}, 2019.

\bibitem[Zhan et~al.(2023)Zhan, Fan, Ye, and Cao]{zhan2023a2s}
Lin Zhan, Jiayuan Fan, Peng Ye, and Jianjian Cao.
\newblock A2s-nas: Asymmetric spectral-spatial neural architecture search for hyperspectral image classification.
\newblock In \emph{ICASSP 2023-2023 IEEE International Conference on Acoustics, Speech and Signal Processing (ICASSP)}, pages 1--5. IEEE, 2023.

\bibitem[Zhang et~al.(2020)Zhang, Li, Pan, Chang, Zhou, Ge, and Su]{zhang2020one}
Miao Zhang, Huiqi Li, Shirui Pan, Xiaojun Chang, Chuan Zhou, Zongyuan Ge, and Steven Su.
\newblock One-shot neural architecture search: Maximising diversity to overcome catastrophic forgetting.
\newblock \emph{IEEE Transactions on Pattern Analysis and Machine Intelligence}, 43\penalty0 (9):\penalty0 2921--2935, 2020.

\bibitem[Zhang et~al.(2021)Zhang, Su, Pan, Chang, Abbasnejad, and Haffari]{zhang2021idarts}
Miao Zhang, Steven~W Su, Shirui Pan, Xiaojun Chang, Ehsan~M Abbasnejad, and Reza Haffari.
\newblock idarts: Differentiable architecture search with stochastic implicit gradients.
\newblock In \emph{International Conference on Machine Learning}, pages 12557--12566. PMLR, 2021.

\bibitem[Zheng et~al.(2022)Zheng, Liu, Lai, and Prakash]{zheng2022coverage}
Haizhong Zheng, Rui Liu, Fan Lai, and Atul Prakash.
\newblock Coverage-centric coreset selection for high pruning rates.
\newblock \emph{arXiv preprint arXiv:2210.15809}, 2022.

\bibitem[Zhong et~al.(2020)Zhong, Yang, Deng, Yan, Wu, Shao, and Liu]{zhong2020blockqnn}
Zhao Zhong, Zichen Yang, Boyang Deng, Junjie Yan, Wei Wu, Jing Shao, and Cheng-Lin Liu.
\newblock Blockqnn: Efficient block-wise neural network architecture generation.
\newblock \emph{IEEE transactions on pattern analysis and machine intelligence}, 43\penalty0 (7):\penalty0 2314--2328, 2020.

\bibitem[Zoph et~al.(2018)Zoph, Vasudevan, Shlens, and Le]{zoph2018learning}
Barret Zoph, Vijay Vasudevan, Jonathon Shlens, and Quoc~V Le.
\newblock Learning transferable architectures for scalable image recognition.
\newblock In \emph{Proceedings of the IEEE conference on computer vision and pattern recognition}, pages 8697--8710, 2018.

\end{thebibliography}
